\documentclass[sigconf]{acmart}
\renewcommand\footnotetextcopyrightpermission[1]{}
\AtBeginDocument{%
  }

\usepackage{graphicx}
\usepackage{booktabs}
\usepackage{multirow}
\usepackage{colortbl}
\usepackage{graphicx}
\definecolor{line-blue}{RGB}{238, 243, 252}

\begin{document}

\title{SpatialImaginer: Towards Adaptive Visual Imagination \\for Spatial Reasoning}

\author{Yian Li$^{1}$, Yang Jiao$^{1}$, Bin Zhu$^{2}$, Tianwen Qian$^{3}$, Shaoxiang Chen$^{4}$,\\ Jingjing Chen$^{5}$, Yu-Gang Jiang$^{5}$}
\email{yali24@m.fudan.edu.cn}
\affiliation{%
  \institution{$^{1}$ College of Computer Science and Artificial Intelligence, Fudan University\\
  $^{2}$ School of Computing and Information Systems, Singapore Management University\\
  $^{3}$ School of Computer Science and Technology, East China Normal University\\
  $^{4}$ MiniMax\\
  $^{5}$ Institute of Trustworthy Embodied Al, Fudan University
  }
  \city{}
  \country{}}

\renewcommand{\shortauthors}{Yian Li et al.}
\renewcommand{\shorttitle}{SpatialImaginer: Towards Adaptive Visual Imagination for Spatial Reasoning}

\begin{abstract}
Spatial intelligence, which refers to the ability to reason about geometric and physical structure from visual observations, remains a core challenge for multimodal large language models. Despite promising performance, recent multimodal large language models (MLLMs) often exhibit fragile reasoning traces in spatial intelligence tasks that involve consistent spatial state recognition.
We argue that these failures stem from a mismatch between the spatial recognition mechanism and the text-only reasoning behavior of these MLLMs. Effective spatial reasoning requires low-level geometric structure to be faithfully preserved and updated throughout the reasoning process, whereas textual representations tend to abstract away precisely these critical details. To address this issue, we propose \textbf{SpatialImaginer}, a unified multimodal generation framework that integrates textual reasoning with visual imagination. Our framework adopts a divide-and-conquer strategy, using text chain-of-thought for high-level semantic planning and the visual imagination for geometry-sensitive state transformation and consistency preservation.
To support this capability, we further introduce a difficulty-aware data engine with closed-loop verification to train the model to invoke visual imagination selectively when stable spatial state tracking is required. Extensive experiments on diverse spatial intelligence benchmarks show that SpatialImaginer achieves state-of-the-art performance and substantially improves robustness on complex multi-step spatial reasoning tasks.
\end{abstract}



\maketitle

\begin{figure}[t]
    \centering
    \vspace{1.5em}
    \includegraphics[width=\linewidth]{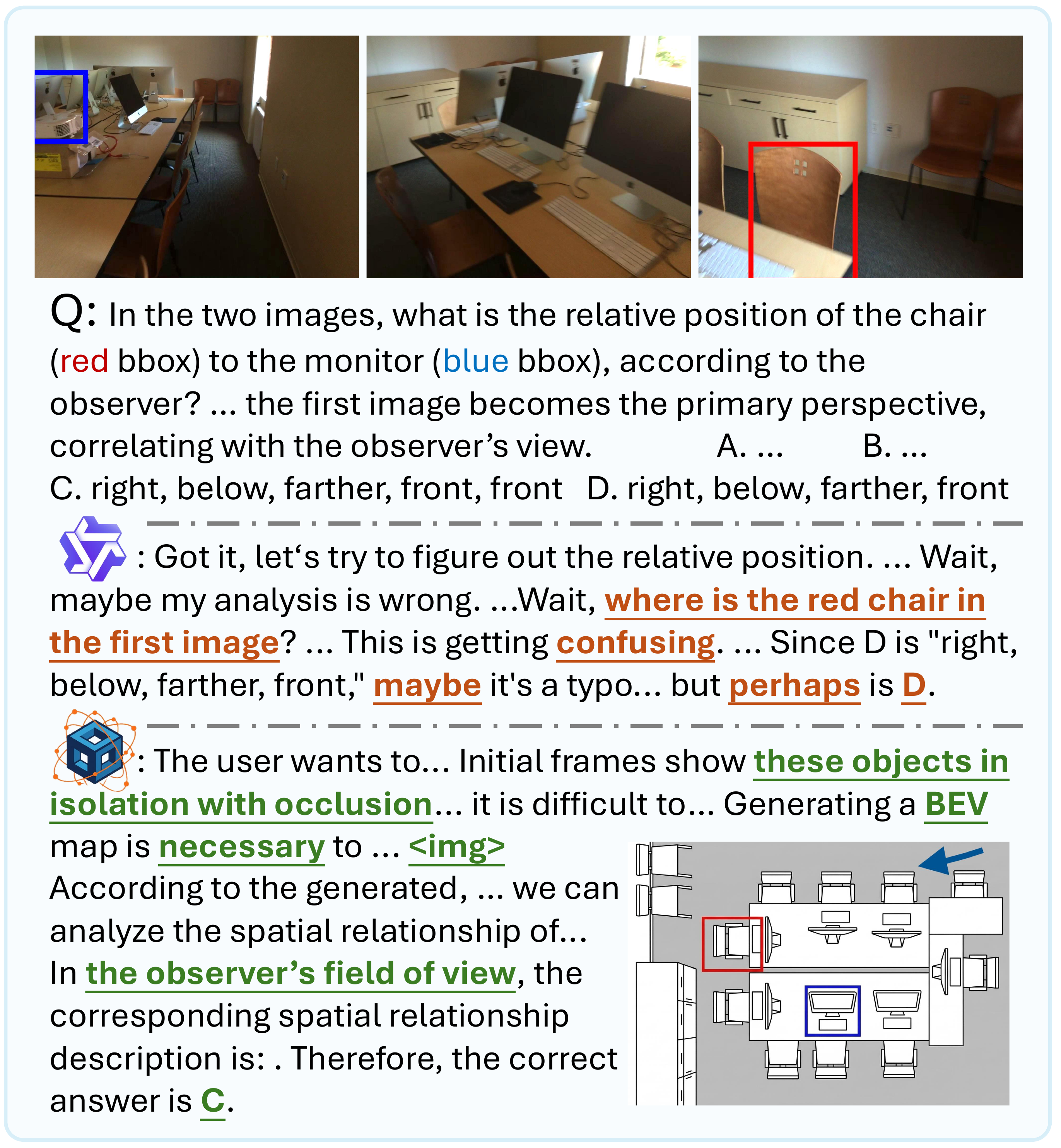}
    \vspace{-1.5em}
    \caption{Unlike text-only reasoning (middle), which often loses geometric consistency across multiple views, SpatialImaginer (bottom) accurately solves complex spatial queries by externalizing intermediate reasoning states into the visual modality.}
    \label{fig:demo}
\end{figure}

\section{Introduction}
\label{sec:intro}

Spatial intelligence, defined as the ability to reason about the geometric and physical structure of the world from visual observations, remains a fundamental challenge for multimodal large language models (MLLMs) and an essential capability for embodied intelligence. Unlike conventional image question answering, which mainly focuses on semantic recognition in static images, spatial intelligence requires the model to maintain a coherent spatial state throughout reasoning. The challenge therefore lies not simply in recognizing scene content, but in preserving spatial consistency as reasoning unfolds. Despite strong performance on a broad range of vision-language tasks, contemporary MLLMs~\cite{Qwen2.5-VL, Qwen3-VL, zhu2025internvl3, deng2025bagel} still exhibit limited robustness in scenarios requiring multi-step spatial reasoning.

A straightforward remedy is to equip MLLMs with textual chain-of-thought (CoT) reasoning, which has proven effective in STEM domains~\cite{wei2022chain, jaech2024openai,li2025look,li2026vigil}. However, these benefits do not readily extend to spatial intelligence. Recent large-scale studies, such as SenseNova-SI~\cite{cai2025scaling}, show that increasing textual reasoning depth yields only modest gains on challenging spatial tasks and can even hurt performance on rigorous benchmarks such as VSI-Bench~\cite{yang2025thinking}. As shown in Figure~\ref{fig:demo}, even advanced MLLMs such as Qwen3-VL~\cite{Qwen3-VL} reveal significant geometric inconsistencies in their text-only reasoning traces when handling spatial intelligence tasks, such as reference frames drift, object correspondences break, and topological relations are misinferred. Therefore, rather than providing the stable scaffold they are intended to offer, these faulty reasoning traces permit errors to compound across steps and ultimately lead the final prediction astray.

We attribute this phenomenon to a fundamental gap between spatial cognition mechanism and text-based representation. Effective spatial reasoning requires the spatial state to be preserved and faithfully updated as reasoning unfolds, which in turn depends on maintaining low-level geometric details, including object correspondence, relative orientation, visibility, and topological consistency. Text, by contrast, serves as a semantic abstraction over raw perceptual signals such as vision, prioritizing high-level concepts over precise geometric structure. Consequently, when spatial information is expressed purely in language, essential geometric details are susceptible to degradation or omission. As reasoning unfolds, these distortions accumulate over time, gradually misleading the reasoning trajectory as demonstrated in Figure~\ref{fig:demo}.

Inspired by the cognitive mechanism of mental rotation~\cite{shepard1988mental}, where humans solve spatial tasks by internally constructing and manipulating visual representations, we propose SpatialImaginer, a unified multimodal generation architecture that externalizes geometry critical intermediate reasoning states into the visual modality.
Specifically, our approach follows a divide-and-conquer design philosophy. Rather than relying on a single modality for spatial reasoning, we assign high-level semantic planning to the textual modality, while entrusting low-level geometric transformations and topological constraints to the visual modality.
The core insight behind this design stems from the heterogeneous nature of queries in spatial intelligence. 
Some queries are primarily semantic, where the main challenge lies in object recognition or attribute discrimination, for which textual reasoning is sufficient. Others are bottlenecked by explicit spatial state tracking, requiring the model to preserve viewpoint changes, geometric transformations, and topological relations across multiple reasoning steps. For such cases, visual reasoning provides a more natural substrate for representing and checking evolving spatial structure, as discussed above.
To enable this selective behavior, we develop an automated data engine with difficulty-aware routing and closed-loop verification. Rather than applying visual imagination uniformly, the engine constructs training signals that encourage the model to rely on text for semantically simple cases while invoking visual imagination only when stable spatial state tracking is necessary. By coupling a unified multimodal generation architecture with this adaptive training mechanism, SpatialImaginer learns to switch flexibly between textual reasoning and visual imagination reasoning according to the demands of the query.

In summary, our contributions are three-fold:
\begin{itemize}
    \item We identify the limitations of text-only chain-of-thought reasoning for spatial intelligence and demonstrate that incorporating visual imagination serves as an effective remedy.

    \item We propose \textbf{SpatialImaginer}, a unified multimodal generation architecture that selectively combines textual and visual reasoning based on the demands of different queries.

    \item Extensive experiments demonstrate that SpatialImaginer achieves state-of-the-art performance across a wide range of spatial intelligence benchmarks.
\end{itemize}

\section{Related Work}
\label{sec:related}

\noindent\textbf{Spatial Intelligence in MLLMs.}
Spatial intelligence is fundamental to embodied AI, yet current MLLMs~\cite{liu2023visual, bai2025qwen3, wang2025internvl3,liu2026enhancing} exhibit persistent limitations when transitioning from 2D visual understanding to dynamic 3D physical spaces. Recent benchmarks like VSI-Bench~\cite{yang2025thinking} systematically evaluate this gap across diverse spatiotemporal tasks, exposing significant deficits in both proprietary~\cite{openai_gpt5_systemcard, gemini_3_pro_systemcard} and open-source models. 
To narrow this gap, early works introduce 3D expert modules to bridge 2D pixels and 3D geometry, such as leveraging pre-trained encoders~\cite{wang2025vggt, zheng2025learning, li2026thinking} in Spatial-MLLM~\cite{wu2025spatial} or explicitly fusing camera parameters in VLM-3R~\cite{fan2025vlm}. However, these hard coded architectures are often constrained by specific sensor configurations and generalize poorly to open domain scenarios. Consequently, recent efforts emphasize large scale data curation and advanced training paradigms. SpatialVLM~\cite{chen2024spatialvlm} pioneered spatial dataset synthesis, followed by SPAR~\cite{zhang2025flatland} and Cambrian-S~\cite{yang2025cambrian}, which aggregate diverse 3D scenes for comprehensive spatial cognition. Beyond data scaling, models like VST~\cite{vst2025}, SpaceR~\cite{ouyang2025spacer}, MindCube~\cite{yin2025spatial(mindcube)}, and SpatialLadder~\cite{li2025spatialladder} explore reinforcement learning and verifiable rewards to optimize structured spatial reasoning. Notably, the comprehensive scaling study SenseNova-SI~\cite{cai2025scaling} reveals that simply scaling text based CoT yields diminishing returns for spatial tasks. Despite these advances, existing methods predominantly rely on textual reasoning, facing a fundamental representation mismatch between discrete linguistic tokens and continuous 3D geometric spaces.

\vspace{1mm}
\noindent\textbf{Multimodal Chain-of-Thought.}
The rapid advancement of unified multimodal architectures~\cite{team2024chameleon, deng2025bagel,jiao2025unitoken} has catalyzed a new paradigm leveraging visual generation to facilitate multimodal understanding. Pioneering works explore innate visual reasoning by generating intermediate images to aid deduction, such as visualizing maze paths~\cite{wu2024mind}, rendering temporal transitions~\cite{li2025imagine}, or drawing 2D auxiliary lines~\cite{li2025zebra, gu2025thinkmorph}. However, visual transformations in these tasks can often be losslessly expressed through text. In contrast, authentic 3D spatial intelligence demands tracking continuous geometric variations, including viewpoint shifts, occlusions, and precise topological relations. These dense physical changes cannot be compressed into discrete text without severe information loss. Addressing this fundamental representation gap, our proposed SpatialImaginer explicitly generates geometrically faithful visual states to natively preserve these complex 3D structures during spatial bottlenecks.

\begin{figure*}[t]
  \centering
  \includegraphics[width=\linewidth]{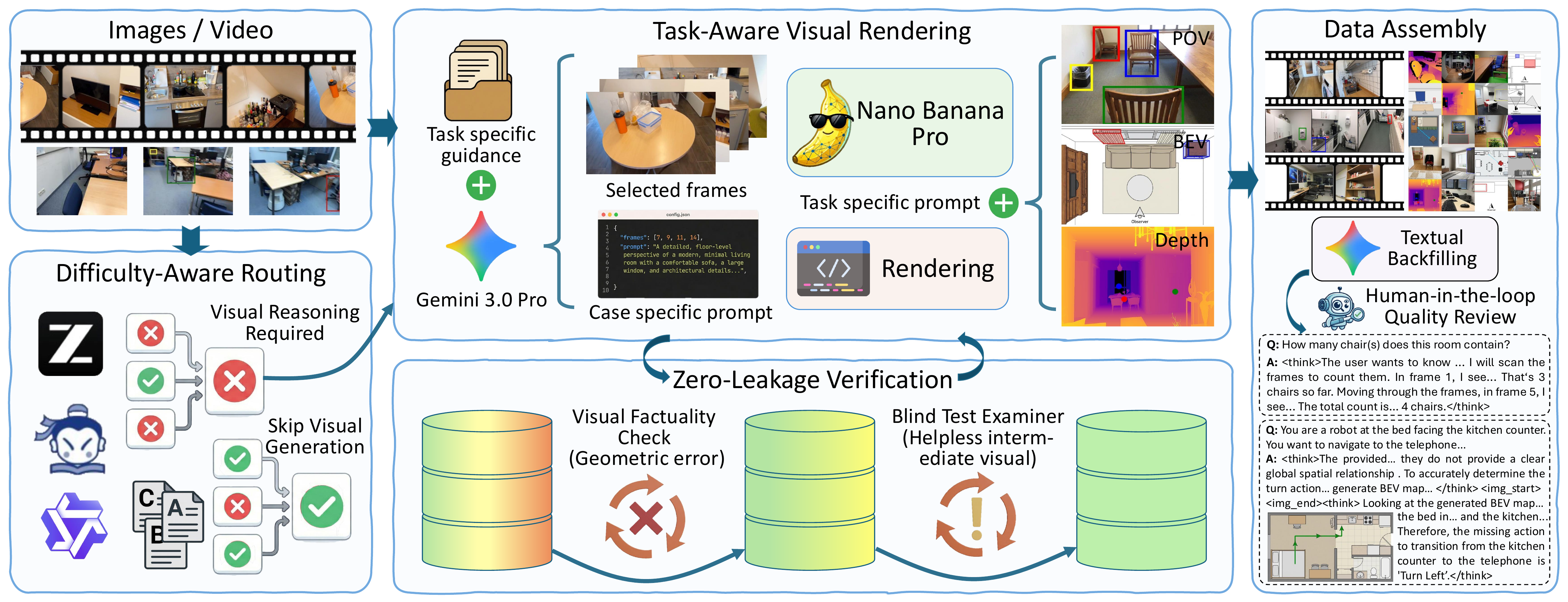}
  \vspace{-2em}
  \caption{Overview of the data engine for interleaved visual reasoning. The pipeline isolates complex spatial bottlenecks via difficulty-aware routing to selectively trigger task-aware visual rendering of intermediate geometric states. Synthesized visuals undergo rigorous zero-leakage verification before being assembled with backfilled textual chains and refined via human-in-the-loop review, yielding an adaptive and balanced multi-modal reasoning dataset.}
  \label{fig:data}
\end{figure*}

\section{Data Engine}
\label{sec:data_engine}

\begin{figure}[t]
    \centering
    \begin{minipage}{0.61\linewidth}
        \centering
        \includegraphics[width=\linewidth]{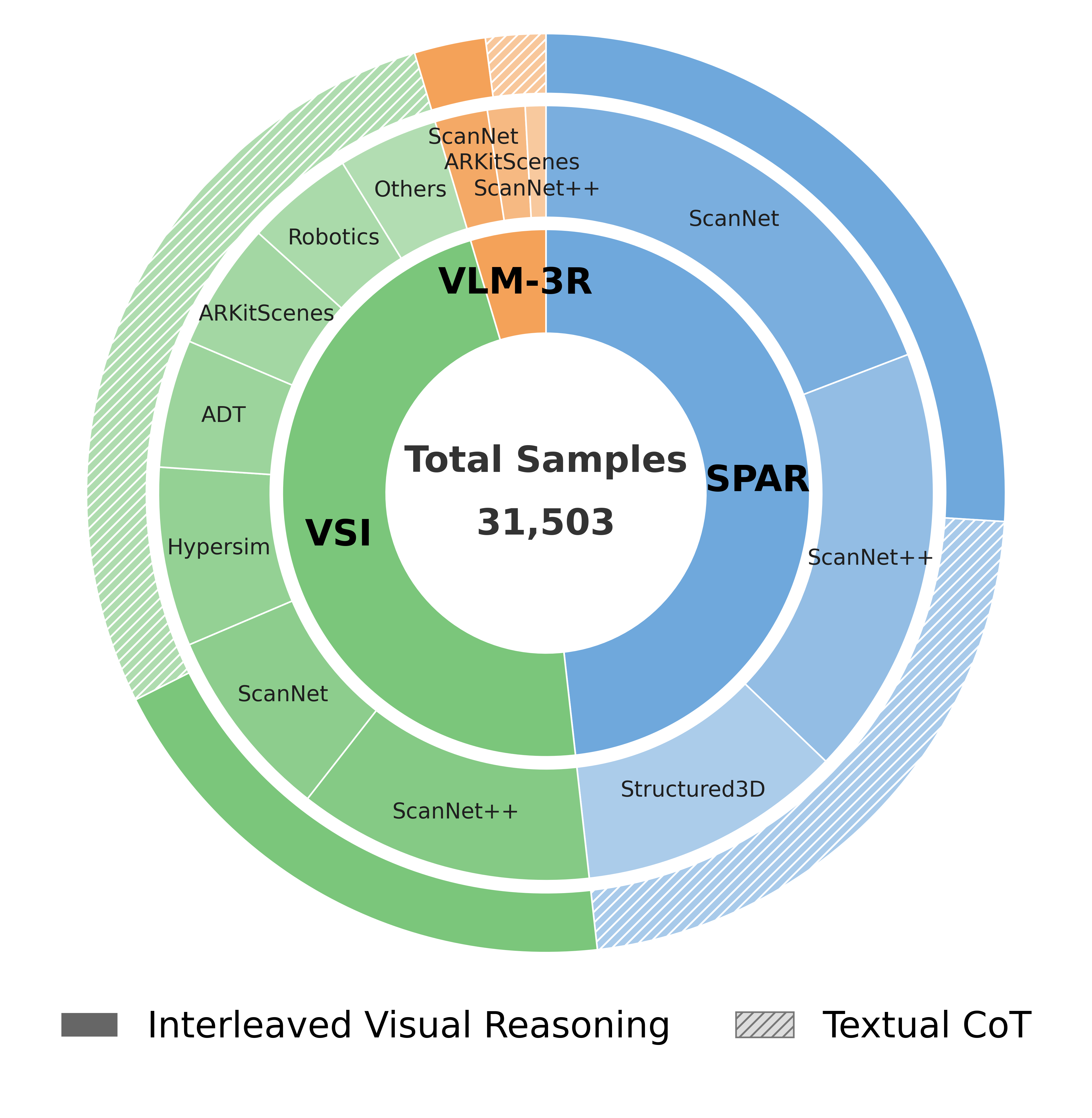}
    \end{minipage}
    \begin{minipage}{0.38\linewidth}
        \centering
        \includegraphics[width=\linewidth]{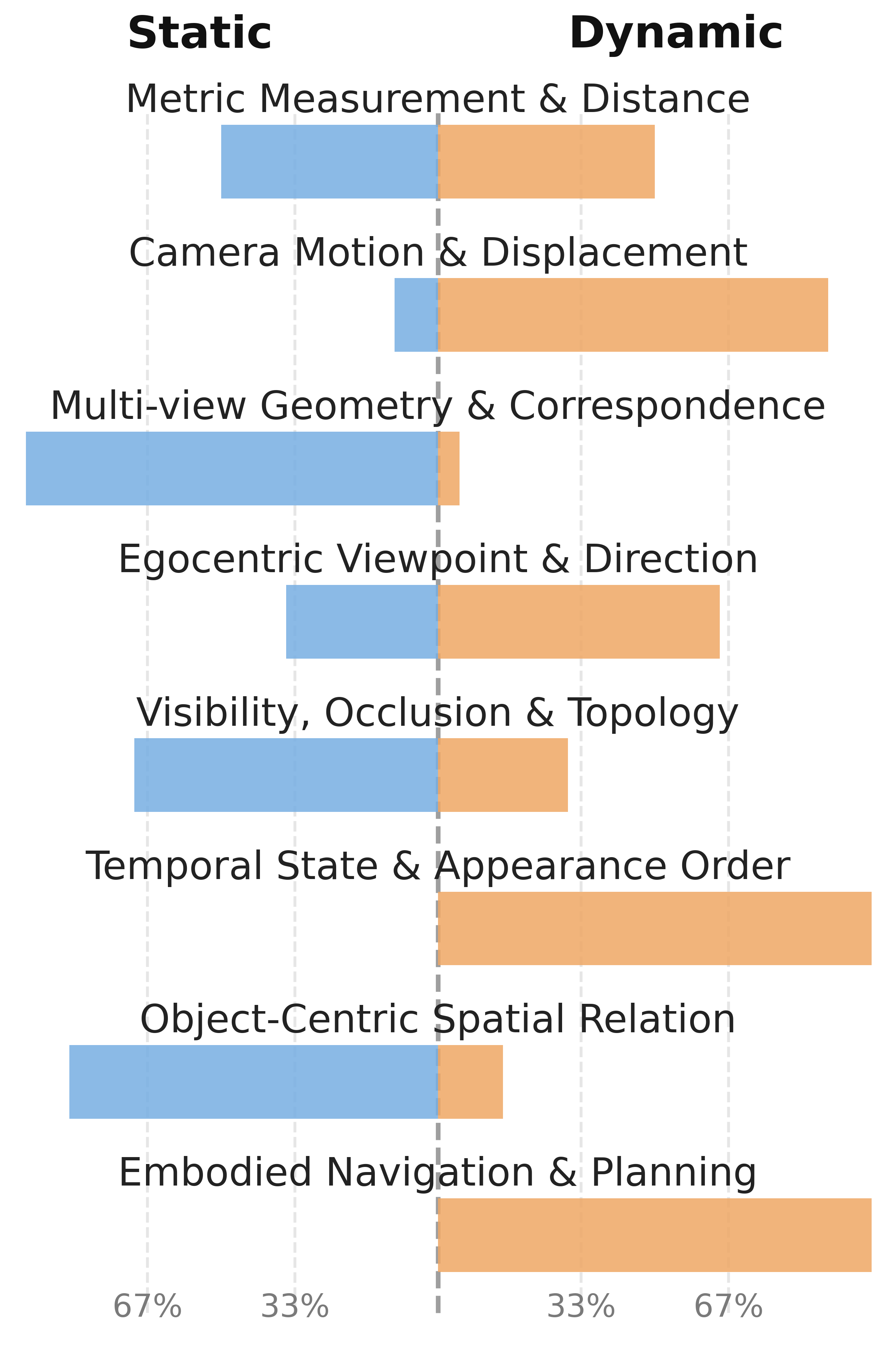}
    \end{minipage}
    \vspace{-1.2em}
    \caption{Statistics of the curated dataset. Left: Hierarchical breakdown of our curated dataset by source domain, environment, and reasoning pattern. Right: Complementary coverage of static and dynamic data across eight core spatial tasks.}
    \label{fig:combined_stat}
\end{figure}

To address the limitations of purely textual spatial reasoning and support our interleaved framework, we design an automated data engine. The pipeline operates over three spatial corpora: SPAR-7M~\cite{zhang2025flatland}, VSI-590K~\cite{yang2025cambrian}, and VLM-3R~\cite{fan2025vlm}. As illustrated in Figure~\ref{fig:data}, the engine integrates dynamic difficulty routing, structured visual rendering, textual backfilling, and closed-loop verification. Instead of unconditionally invoking generative modules, this framework selectively synthesizes intermediate visual states only for complex spatial bottlenecks, yielding an adaptive dataset that aligns visual generation with task difficulty.

\subsection{Difficulty-Aware Routing}
Spatial reasoning tasks exhibit substantial variation in complexity. While basic geometric queries can often be resolved via textual Chain of Thought, complex kinematic simulations or multi-step perspective transformations are prone to reference frame drift and topological errors when constrained to discrete token sequences. 

To optimize computation and reduce noise, we introduce a difficulty-aware routing mechanism. Specifically, we deploy three advanced VLM, including Qwen3-VL~\cite{Qwen3-VL}, InternVL3~\cite{zhu2025internvl3} and GLM-4.1V~\cite{hong2025glm}, as a front end prober to execute three independent inference runs on each candidate query. A majority voting rejection protocol is applied: a query is routed to the visual generation engine only if the baseline model fails consistently (e.g., incorrect on at least two out of the three attempts). This feedback driven mechanism serves as an empirical decision boundary, isolating difficult samples that exceed the capacity of purely textual spatial modeling and strictly require explicit geometric grounding.

\subsection{Task-Oriented Rendering and Verification}
For queries routed to this phase, the engine synthesizes an intermediate visual state ($x_{mid}$) to provide explicit geometric grounding for subsequent deductive reasoning. Crucially, $x_{mid}$ is designed as a structured explanatory representation rather than an unconstrained generation. To achieve high-fidelity synthesis, we develop a task-aware rendering pipeline equipped with customized prompting and two stage verification.

\vspace{1mm}
\noindent\textbf{Dynamic Keyframe Extraction.} 
For dynamic environments involving continuous video sequences, processing dense frames in their entirety is computationally prohibitive. We employ an advanced MLLM (Gemini-3-Pro) to actively extract a minimal subset of informative keyframes. Conditioned on these extracted frames, the original spatial query, and the ground truth answer, the model formulates customized generation prompts tailored to specific task categories. This explicit conditioning steers the generation toward structured geometric evidence, reducing the model's tendency to produce unconstrained or task irrelevant visual artifacts.

\vspace{1mm}
\noindent\textbf{Task-Specific Modality Routing.} 
To reflect distinct spatial constraints, the rendering process is routed based on physical modalities. For global topology queries like route planning tasks and egocentric perspective shifts, the rendering process is explicitly conditioned on task specific spatial modalities using Nano Banana Pro (gemini-3-pro-image-preview) to synthesize 2D Bird's-Eye-View (BEV) maps or Novel Point-of-View (POV) images, respectively. Both scenarios enforce the preservation of original object markers and spatial layouts. Alternatively, for camera centric distance queries where generative models typically struggle with metric depth, we bypass generative synthesis. Instead, we deterministically project raw ground truth depth maps into pseudo-color visualizations overlaid with bounding boxes, ensuring metrically accurate spatial representations.

\vspace{1mm}
\noindent\textbf{Zero-Leakage Policy \& Verification.} 
To ensure both structural correctness and functional utility, synthesized images undergo a two step verification process. During rendering, prompts explicitly prevent the inclusion of textual answers or numerical measurements within $x_{mid}$. To ensure structural correctness and functional utility, the synthesized images undergo a rigorous two step verification process. In the first phase, Gemini-3-Pro conducts a \textit{visual factuality check} to filter out images with obvious structural errors or topology corruption. In the second phase, Qwen3-VL-8B acts as a \textit{blind test examiner} to evaluate whether $x_{mid}$ actually facilitates spatial comprehension. We prompt this examiner to answer the original query conditioned solely on the generated $x_{mid}$. If the model fails to reach the ground truth answer, the intermediate state is discarded as ineffectual. Through this filtering pipeline, approximately 48.2\% of the generated candidates are retained.

\subsection{Textual Backfilling and Formulation}
Samples that successfully pass the verification process are structured by Gemini-3-Pro into a standardized interleaved format: spatial planning $\rightarrow$ intermediate visual  ($x_{mid}$) $\rightarrow$ deductive resolution. However, curating a dataset exclusively composed of interleaved visual proofs introduces a spurious inductive bias. It conditions the model toward obligatory visual generation, forcing it to render intermediate images regardless of the actual task complexity. 

To counteract this bias and enforce adaptive reasoning patterns, we introduce a complementary textual backfilling strategy. We retrieve candidate queries from the initial routing stage that are reliably resolvable via purely textual logic, and utilize Gemini-3-Pro to synthesize high quality textual reasoning chains for them. After a human-in-the-loop quality review to filter flawed annotations, we assemble a balanced distribution of both interleaved and pure textual samples. We equip the model to autonomously assess task complexity during the planning phase.

To seamlessly unify these modalities within our dataset, each assembled sequence is formatted as a structured tuple:
\begin{equation}
    \mathcal{C} = \{C_{plan}, C_{vis}, C_{deduct}\}
\end{equation}
where $C_{plan}$ and $C_{deduct}$ represent the textual reasoning segments, and $C_{vis}$ encapsulates the intermediate visual state ($x_{mid}$). Crucially, for the backfilled purely textual samples, this visual component simply reduces to an empty set ($C_{vis} = \emptyset$). This standardized data formulation aligns both reasoning paths into a single distribution, explicitly preparing the model to learn adaptive mode-switching during the subsequent optimization phase.

\subsection{Dataset Composition and Statistics}
Using the proposed engine, we curate a spatial intelligence dataset comprising 31,503 samples. The dataset exhibits a balanced reasoning distribution: 15,077 samples (47.86\%) are formatted as Interleaved Visual CoT, while 16,426 samples (52.14\%) utilize purely textual deduction. To ensure robust generalization, the dataset is assembled from complementary source domains:

\vspace{1mm}
\noindent\textbf{SPAR Subset (48.21\%).} 
Comprising 15,189 samples sourced from the SPAR~\cite{zhang2025flatland} dataset (which aggregates ScanNet~\cite{dai2017scannet}, ScanNet++~\cite{yeshwanth2023scannetpp}, and Structured3D~\cite{Structured3D}), this subset covers 171 spatial reasoning categories. It predominantly utilizes static 1-to-3 multi-view image inputs (89.3\%), fostering deep geometric deduction across diverse static spatial configurations.

\vspace{1mm}
\noindent\textbf{VSI \& VLM-3R Subset (51.79\%).} 
Comprising 16,314 samples derived from the VSI-590K~\cite{yang2025cambrian} corpus (aggregating diverse environments such as Hypersim~\cite{roberts2021hypersim}, ScanNet++~\cite{yeshwanth2023scannetpp}, ProcTHOR~\cite{deitke2022ProcTHOR}, and real world videos~\cite{pan2023aria, dehghan2021arkitscenes, armeni20163d, dai2017scannet}) and augmented with embodied route planning trajectories from VLM-3R~\cite{fan2025vlm}. This subset focuses on 13 core spatiotemporal capabilities, emphasizing dynamic perception via 16 frame video inputs (73.54\%) to explicitly evaluate the model's ability to maintain spatial coherence over continuous temporal sequences.

\begin{figure*}[!ht]
  \centering
  \includegraphics[width=\linewidth]{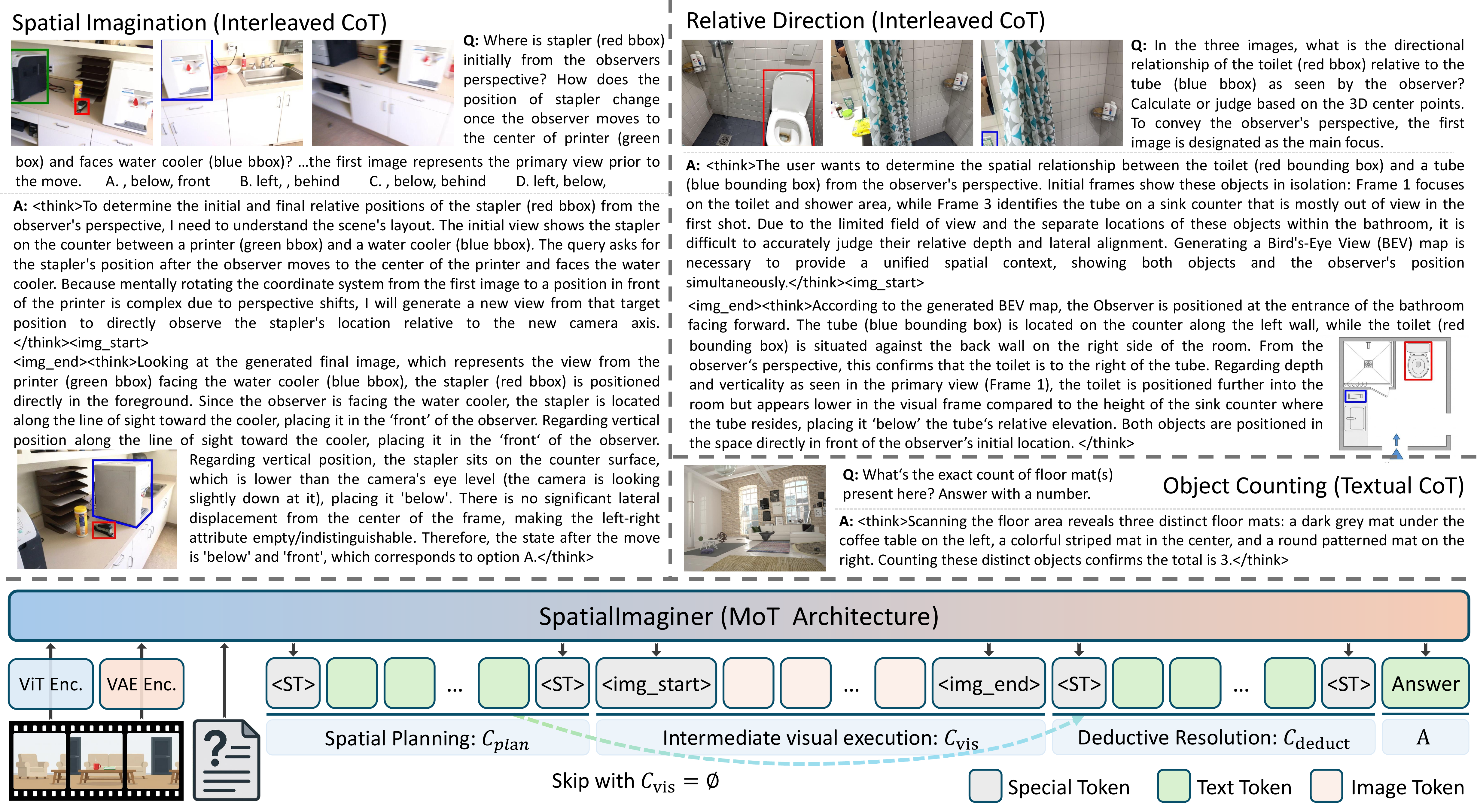}
  \vspace{-2em}
  \caption{The architecture and reasoning paradigms of SpatialImaginer. The model employs a unified framework (bottom) that structures reasoning into spatial planning ($C_{plan}$), intermediate visual execution ($C_{vis}$), and deductive resolution ($C_{deduct}$). As demonstrated in the top examples, the architecture adaptively switches between interleaved multimodal CoT for complex spatial queries (generating visual intermediate states) and purely textual CoT ($C_{vis} = \emptyset$) for simpler semantic tasks.}
  \label{fig:method}
\end{figure*}

\section{Interleaved Visual Reasoning Framework}
\label{sec:method_training}

To operationalize the interleaved spatial dataset curated by our engine described in Section \ref{sec:data_engine}, we introduce the architecture and optimization strategy of SpatialImaginer. The primary objective of this framework is to seamlessly unify continuous visual generation with discrete textual deduction. This unified formulation not only ensures that the model maintains strict geometric consistency during complex spatial transformations but also naturally fosters an adaptive mode switching capability. Consequently, the model learns to autonomously alternate between explicit visual simulation and efficient purely textual reasoning based on the inherent spatial complexity of the query.

\subsection{Unified Architecture and Problem Formulation}
We instantiate our framework upon the unified multimodal architecture of BAGEL~\cite{deng2025bagel}, which natively integrates visual understanding and generation within a single autoregressive parameter space. An input image is encoded into a dual token representation $H_v = [F_v, Z_v]$. Here, $F_v$ denotes the continuous semantic features extracted by a Vision Transformer encoder, while $Z_v$ represents a compact latent space encoding structured visual details via a continuous time VAE. To enable efficient multimodal modeling, the large language model backbone employs a hybrid attention mechanism: textual tokens utilize standard causal masking, whereas bidirectional attention is applied internally within the visual tokens. Furthermore, special delimiter tokens such as \texttt{<img\_start>} and \texttt{<img\_end>} are introduced to explicitly mark the boundaries of visual segments. These specific tags provide a definitive structure that enables the seamless integration of continuous visual features and discrete textual elements within a unified reasoning sequence.

Given an initial visual observation $X$ and a spatial query $Q$, SpatialImaginer does not directly regress the final answer $A$. Instead, it autoregressively formulates the structured interleaved sequence $\mathcal{C} = (C_{plan}, C_{vis}, C_{deduct})$ defined in our data engine. The generation trajectory is probabilistically decomposed as:
\begin{equation}
\begin{aligned}
P(A, \mathcal{C} | X, Q) &= \underbrace{P(C_{plan} | X, Q)}_{\text{Spatial Planning}} \cdot \underbrace{P(C_{vis} | C_{plan}, X, Q)}_{\text{Optional Visual Execution}} \\
&\quad \cdot \underbrace{\prod_{k=1}^{L} P(a_k | a_{<k}, C_{vis}, C_{plan}, X, Q)}_{\text{Deductive Resolution \& Answer}}
\end{aligned}
\end{equation}
where $a_k$ denotes the discrete textual tokens comprising both the deductive resolution ($C_{deduct}$) and the final answer $A$. 

This explicit formulation serves a critical cognitive purpose. When relying purely on textual deduction, models must track complex spatial relationships entirely through language. Without explicit physical constraints, this process frequently loses spatial coherence during complex transformations. By predicting the intermediate visual state $C_{vis}$, the model translates implicit spatial operations into an observable representation such as a depth map or a bird's eye view. This generated image functions as a tangible geometric anchor. It visually grounds the subsequent textual deduction $C_{deduct}$, ensuring that the logical reasoning adheres to actual physical layouts rather than producing unsupported assumptions.

\begin{table*}[ht]
  \begin{center}
        \setlength{\tabcolsep}{1.8pt}
        \resizebox{\textwidth}{!}{%
        \begin{tabular}{l|c|cccc|ccc|ccccc}
          \toprule
          \multirow{2}{*}[-0.6ex]{Methods} & \multirow{2}{*}[-0.6ex]{Avg.}
          & \multicolumn{4}{c|}{\cellcolor{green!10}Low}
          & \multicolumn{3}{c|}{\cellcolor{yellow!10}Mid}
          & \multicolumn{5}{c}{\cellcolor{orange!10}High} \\
          \cmidrule(lr){3-6}\cmidrule(lr){7-9}\cmidrule(lr){10-14}
          &  & Depth-OC & Depth-OO & Dist-OC & Dist-OO & PosMatch & CamPose & ViewChgI & DistI-OO & ObjRel-OC & ObjRel-OO & SpImag-OC & SpImag-OO \\
          \midrule
          \rowcolor{black!9}\multicolumn{14}{l}{\textbf{Proprietary Models}} \\
          Seed-1.6~\cite{seed2025seed1_5vl} & 50.1 & -- & -- & -- & -- & -- & -- & -- & -- & -- & -- & -- & -- \\
          Gemini-3-Pro~\cite{gemini_3_pro_systemcard} & 48.7 & 45.1 & 23.2 & 53.3 & 42.0 & 61.6 & 25.8 & 23.2 & 77.7 & 61.3 & 61.7 & 55.1 & 43.8 \\
          Grok-4~\cite{grok4_xai_2025} & 44.8 & -- & -- & -- & -- & -- & -- & -- & -- & -- & -- & -- & -- \\
          GPT-5~\cite{openai_gpt5_systemcard} & 49.7 & 53.1 & 22.8 & 52.0 & 42.3 & 65.1 & 24.5 & 27.7 & 74.9 & 64.5 & 64.3 & 49.7 & 47.6 \\
          \midrule
          \rowcolor{black!9}\multicolumn{14}{l}{\textbf{Open-source General Models}} \\
          Bagel-7B-MoT~\cite{deng2025bagel} & 39.1 & 43.3 & 20.9 & 53.1 & 21.0 & 58.8 & 32.5 & 15.1 & 57.7 & 51.5 & 46.7 & 37.0 & 32.5 \\
          Qwen2.5-VL-7B-Instruct~\cite{Qwen2.5-VL} & 33.8 & 33.6 & 18.1 & 35.6 & 23.2 & 41.2 & 23.8 & 17.7 & 56.7 & 44.3 & 43.6 & 33.0 & 31.2 \\
          Qwen3-VL-8B-Instruct~\cite{Qwen3-VL} & 39.6 & 48.7 & 19.9 & 26.9 & 46.9 & 52.2 & 26.0 & 16.1 & 69.6 & 50.5 & 49.9 & 30.2 & 32.0 \\
          InternVL3-8B~\cite{zhu2025internvl3} & 35.9 & 25.5 & 21.2 & 16.8 & 38.0 & 62.9 & 36.3 & 18.8 & 63.2 & 40.8 & 46.5 & 35.5 & 32.9 \\
          \midrule
          \rowcolor{black!9}\multicolumn{14}{l}{\textbf{Open-source Spatial Intelligence Models}} \\
          MindCube-3B~\cite{yin2025spatial(mindcube)} & 20.8 & 22.0 & 15.3 & 13.0 & 16.7 & 22.4 & 20.5 & 25.0 & 23.2 & 24.0 & 24.3 & 24.4 & 23.4 \\
          SpatialLadder-3B~\cite{li2025spatialladder} & 32.9 & 26.7 & 23.0 & 16.0 & 33.8 & 46.3 & 28.3 & 13.6 & 55.2 & 51.8 & 39.2 & 36.8 & 28.9 \\
          Spatial-MLLM-4B~\cite{wu2025spatial} & 35.3 & -- & -- & -- & -- & -- & -- & -- & -- & -- & -- & -- & -- \\
          SpaceR-7B~\cite{ouyang2025spacer} & 34.2 & 37.9 & 16.1 & 23.1 & 30.8 & 34.1 & 20.8 & 16.7 & 54.0 & 49.5 & 47.6 & 34.3 & 37.8 \\
          VLM-3R~\cite{fan2025vlm} & 42.4 & -- & -- & -- & -- & -- & -- & -- & -- & -- & -- & -- & -- \\
          ViLaSR-7B~\cite{wu2025reinforcing(vilasr)} & 37.4 & 41.0 & 18.7 & 45.9 & 35.7 & 44.0 & 25.3 & 17.8 & 58.3 & 46.8 & 43.5 & 34.9 & 29.0 \\
          VST-7B-SFT~\cite{vst2025} & 46.6 & 59.7 & 26.7 & 42.6 & 46.4 & 35.9 & 31.8 & 16.6 & \textbf{83.9} & 56.3 & 57.4 & 43.5 & 36.1 \\
          Cambrian-S-7B~\cite{yang2025cambrian} & 37.9 & 42.9 & 16.8 & 23.5 & 35.8 & 38.2 & 33.8 & 21.9 & 52.3 & 69.3 & 46.0 & 45.3 & 35.2 \\
          \midrule
          \rowcolor{line-blue}\multicolumn{14}{l}{\textbf{Ours}} \\
          SpatialImaginer\textsubscript{Bagel-7B-MoT} & \textbf{62.3} & \textbf{70.2} & \textbf{35.2} & \textbf{72.8} & \textbf{50.0} & \textbf{84.0} & \textbf{36.0} & \textbf{29.9} & 76.3 & \textbf{84.5} & \textbf{83.5} & \textbf{58.1} & \textbf{60.3} \\
          \bottomrule
        \end{tabular}%
        }
  \end{center}
  \caption{
  \textbf{Comparison on SPAR-Bench~\cite{zhang2025flatland}} across 12 spatial categories stratified by difficulty (Low / Mid / High). All results follow the EASI protocol~\cite{easi2025}. Best open-source results are \textbf{bolded}.
  }
  \vspace{-1.5em}
  \label{tab:spar-bench}
\end{table*}

\subsection{Optimization for Interleaved Reasoning}
To ensure the generated intermediate states $C_{vis}$ serve as reliable geometric proofs rather than unconstrained artifacts, we optimize the unified model through a joint learning objective, preceded by a necessary perception initialization.

\vspace{1mm}
\noindent\textbf{Geometric-Aware Warm-up.} 
Since VAE tokens ($Z_v$) are inherently optimized to reconstruct 2D image textures, exposing them to complex spatial generation tasks without prior 3D metric alignment often results in geometrically inconsistent visual artifacts. To prevent this, we introduce a preliminary geometric-aware warm-up training. During this phase, we temporarily deactivate the VAE generation module and optimize the model using a perception-focused subset of our spatial data, targeting fundamental geometric properties such as metric depth and relative orientation. By feeding only the ViT tokens ($F_v$) into the large language model decoder, we optimize the standard causal language modeling objective:
\begin{equation}
    \mathcal{L}_{\text{ground}} = \mathbb{E}_{(X, Q, A) \sim \mathcal{D}_{\text{perc}}} \left[ -\sum_{i} \log P_{\theta}(a_i | a_{<i}, F_v(X), Q) \right]
\end{equation}
This straightforward initialization instills essential 3D metric priors into the cross-modal projector and the LLM backbone, establishing a structural foundation that guides the subsequent generation phase.

\vspace{1mm}
\noindent\textbf{Joint Interleaved Optimization.} 
Building upon these grounded spatial priors, the primary optimization phase activates both ViT and VAE tokens $[F_v, Z_v]$ to enable closed-loop visual reasoning. We utilize the difficulty stratified interleaved dataset curated by our automated engine. The model is optimized jointly across modalities. For all textual reasoning tokens $C_{plan}$, $C_{deduct}$, and $A$, we apply the standard cross entropy loss $\mathcal{L}_{\text{CE}}$. For the intermediate visual states $C_{vis}$, we employ a Rectified Flow matching objective $\mathcal{L}_{\text{img}}$ to optimize the velocity field of the VAE latent tokens. Let $z_1$ represent the ground truth visual latent and $z_0 \sim \mathcal{N}(0, I)$ denote the initial noise. The visual generation loss is defined as:
\begin{equation}
    \mathcal{L}_{\text{img}} = \mathbb{E}_{t, z_0, z_1} \left\| v_\theta(z_t, t, c) - (z_1 - z_0) \right\|_2^2
\end{equation}
where $z_t = t z_1 + (1-t) z_0$, $v_\theta$ is the predicted velocity field, and $c$ encapsulates the preceding multimodal context. Standard rectified flow samples the timestep $t$ uniformly. However, because generating explicit spatial proofs such as BEV maps demands rigorous global geometric alignment rather than mere texture synthesis, we apply a time shift schedule $\mu=3.0$ to skew the timestep sampling toward the high noise regime. This forces the model to allocate more learning capacity to the crucial structural formation phase. The overall joint objective combines these modalities using balancing coefficients $\lambda_{\text{text}}$ and $\lambda_{\text{img}}$:
\begin{equation}
    \mathcal{L}_{\text{joint}} = \lambda_{\text{text}} \mathcal{L}_{\text{CE}}(C_{plan}, C_{deduct}, A) + \lambda_{\text{img}} \mathcal{L}_{\text{img}}
\end{equation}

Crucially, optimizing this joint objective over our balanced dataset naturally facilitates the emergence of adaptive mode switching. For straightforward semantic queries, the model learns that predicting an empty visual set $C_{vis} = \emptyset$ and relying solely on textual deduction is the most efficient path to minimize the overall training loss. Conversely, for complex spatial bottlenecks, purely textual reasoning often leads to incorrect answers and high cross entropy penalties due to geometric drift. Driven by this natural optimization dynamic, the model autonomously learns to invoke explicit visual generation primarily for challenging scenarios where a geometric anchor improves the final reasoning accuracy.

\begin{table*}[ht]
  \begin{center}
        \resizebox{.97\textwidth}{!}{%
        \begin{tabular}{l|c|cccc|cccc}
          \toprule
          \multirow{2}{*}[-0.6ex]{Methods} & \multirow{2}{*}[-0.6ex]{Avg.}
          & \multicolumn{4}{c|}{\cellcolor{orange!10}Numerical Answer}
          & \multicolumn{4}{c}{\cellcolor{yellow!10}Multiple-Choice Answer} \\
          \cmidrule(lr){3-6}\cmidrule(lr){7-10}
          &  & Obj. Count & Abs. Dist. & Obj. Size & Room Size & Rel. Dist. & Rel. Dir. & Route Plan & Appr. Order \\
          \midrule
          Human & 79.2 & 94.3 & 47.0 & 60.4 & 45.9 & 94.7 & 95.8 & 95.8 & 100.0 \\
          Random Choice(Frequency) & 34.0 & 62.1 & 32.0 & 29.9 & 33.1 & 25.1 & 47.9 & 28.4 & 25.2 \\
          \midrule
          \rowcolor{black!9}\multicolumn{10}{l}{\textbf{Proprietary Models}} \\
          Seed-1.6~\cite{seed2025seed1_5vl} & 49.9 & 43.5 & 34.4 & 66.1 & 52.8 & 55.1 & 35.7 & 44.3 & 68.0 \\
          Gemini-3-Pro~\cite{gemini_3_pro_systemcard} & 52.5 & 38.0 & 37.8 & 72.7 & 44.1 & 59.9 & 55.7 & 45.9 & 66.0 \\
          Grok-4~\cite{grok4_xai_2025} & 47.9 & 37.2 & 33.0 & 60.8 & 45.4 & 53.1 & 39.7 & 47.4 & 66.8 \\
          GPT-5~\cite{openai_gpt5_systemcard} & 55.0 & 53.3 & 34.5 & 73.3 & 47.5 & 63.7 & 48.7 & 50.3 & 68.9 \\
          \midrule
          \rowcolor{black!9}\multicolumn{10}{l}{\textbf{Open-source General Models}} \\
          Bagel-7B-MoT~\cite{deng2025bagel} & 31.4 & 30.1 & 29.2 & 35.5 & 25.8 & 34.9 & 41.4 & 30.4 & 24.1 \\
          Qwen2.5-VL-7B-Instruct~\cite{Qwen2.5-VL} & 32.3 & 32.9 & 18.2 & 43.9 & 31.7 & 38.0 & 37.4 & 28.4 & 28.0 \\
          Qwen3-VL-8B-Instruct~\cite{Qwen3-VL} & 57.9 & 67.6 & 47.0 & 76.3 & 61.9 & 58.0 & 51.0 & 35.0 & 66.3 \\
          InternVL3-8B~\cite{zhu2025internvl3} & 42.1 & 66.0 & 34.9 & 43.6 & 47.5 & 48.0 & 39.3 & 26.3 & 31.4 \\
          \midrule
          \rowcolor{black!9}\multicolumn{10}{l}{\textbf{Open-source Spatial Intelligence Models}} \\
          MindCube-3B~\cite{yin2025spatial(mindcube)} & 17.2 & 12.9 & 22.8 & 4.3 & 23.5 & 20.3 & 15.7 & 16.0 & 22.5 \\
          SpatialLadder-3B~\cite{li2025spatialladder} & 44.9 & 62.2 & 35.4 & 62.0 & 41.4 & 45.6 & 46.5 & 27.3 & 38.5 \\
          Spatial-MLLM-4B~\cite{wu2025spatial} & 46.3 & 66.7 & 38.1 & 63.6 & 35.5 & 40.4 & 48.2 & 33.0 & 44.3 \\
          SpaceR-7B~\cite{ouyang2025spacer} & 41.5 & 44.5 & 24.7 & 53.5 & 37.3 & 41.9 & 46.1 & 29.3 & 54.8 \\
          VLM-3R-7B~\cite{fan2025vlm} & 60.9 & 70.2 & \textbf{49.4} & 69.2 & 67.1 & 65.4 & \textbf{80.5} & 45.4 & 40.1 \\
          ViLaSR-7B~\cite{wu2025reinforcing(vilasr)} & 44.6 & 58.1 & 33.9 & 61.4 & 28.9 & 45.1 & 46.5 & 29.9 & 53.2 \\
          VST-7B-SFT~\cite{vst2025} & 55.5 & 68.8 & 37.3 & \textbf{74.5} & 62.2 & 55.2 & 48.7 & 41.8 & 55.5 \\
          Cambrian-S-7B~\cite{yang2025cambrian} & 62.9 & 68.2 & 45.8 & 72.5 & \textbf{67.6} & 66.8 & 69.6 & 39.2 & \textbf{73.8} \\
          \midrule
          \rowcolor{line-blue}\multicolumn{10}{l}{\textbf{Ours}} \\
          SpatialImaginer\textsubscript{Bagel-7B-MoT} & \textbf{63.7} & \textbf{70.6} & 46.8 & 72.7 & 64.6 & \textbf{67.4} & 69.9 & \textbf{46.4} & 71.2 \\
          \bottomrule
        \end{tabular}%
        }
  \end{center}
  \caption{
  \textbf{Comparison on VSI-Bench~\cite{yang2025thinking}.} Numerical answers are evaluated by MRA and multiple-choice answers by accuracy. All results follow the EASI protocol~\cite{easi2025}. Best open-source results are \textbf{bolded}.
  }
  \vspace{-2em}
  \label{tab:vsi-bench}
\end{table*}

\begin{table*}[ht]
    \centering
    \begin{minipage}{0.75\textwidth}
        \centering
        \vspace{-1em}
          \begin{center}
      \setlength{\tabcolsep}{2pt}
      \resizebox{\textwidth}{!}{%
      \begin{tabular}{l|cccccc}
        \toprule
        Methods & VSI-Bench & SPAR-Bench & ViewSpatial & \hspace{.2em} BLINK \hspace{.2em} & 3DSRBench & EmbSpatial \\
        \midrule
        VLM-3R-7B~\cite{fan2025vlm} & 60.7 & 42.4 & 40.5 & 52.3 & 51.5 & 68.2 \\
        MindCube-3B~\cite{yin2025spatial(mindcube)} & 17.2 & 20.8 & 24.1 & 35.1 & 2.8 & 37.0 \\
        SpatialLadder-3B~\cite{li2025spatialladder} & 44.9 & 32.9 & 39.9 & 43.0 & 42.8 & 58.2 \\
        Spatial-MLLM-4B~\cite{wu2025spatial} & 46.3 & 35.3 & 34.7 & 40.5 & 36.2 & 50.0 \\
        SpaceR-7B~\cite{ouyang2025spacer} & 41.6 & 34.2 & 35.9 & 49.6 & 40.5 & 66.9 \\
        ViLaSR-7B~\cite{wu2025reinforcing(vilasr)} & 44.6 & 37.4 & 35.7 & 51.4 & 46.6 & 67.3 \\
        Cambrian-S-7B~\cite{yang2025cambrian} & 62.9 & 37.9 & 41.3 & 37.9 & 45.0 & \textbf{72.8} \\
        \midrule
        \rowcolor{black!9} \textbf{SpatialImaginer (Ours)} & \textbf{63.7} & \textbf{62.3} & \textbf{44.4} & \textbf{59.8} & \textbf{54.3} & 68.9 \\
        \bottomrule
      \end{tabular}%
      }
  \end{center}

    \end{minipage}
    \hfill
    \begin{minipage}{0.23\textwidth}
        \centering
        \includegraphics[width=\textwidth]{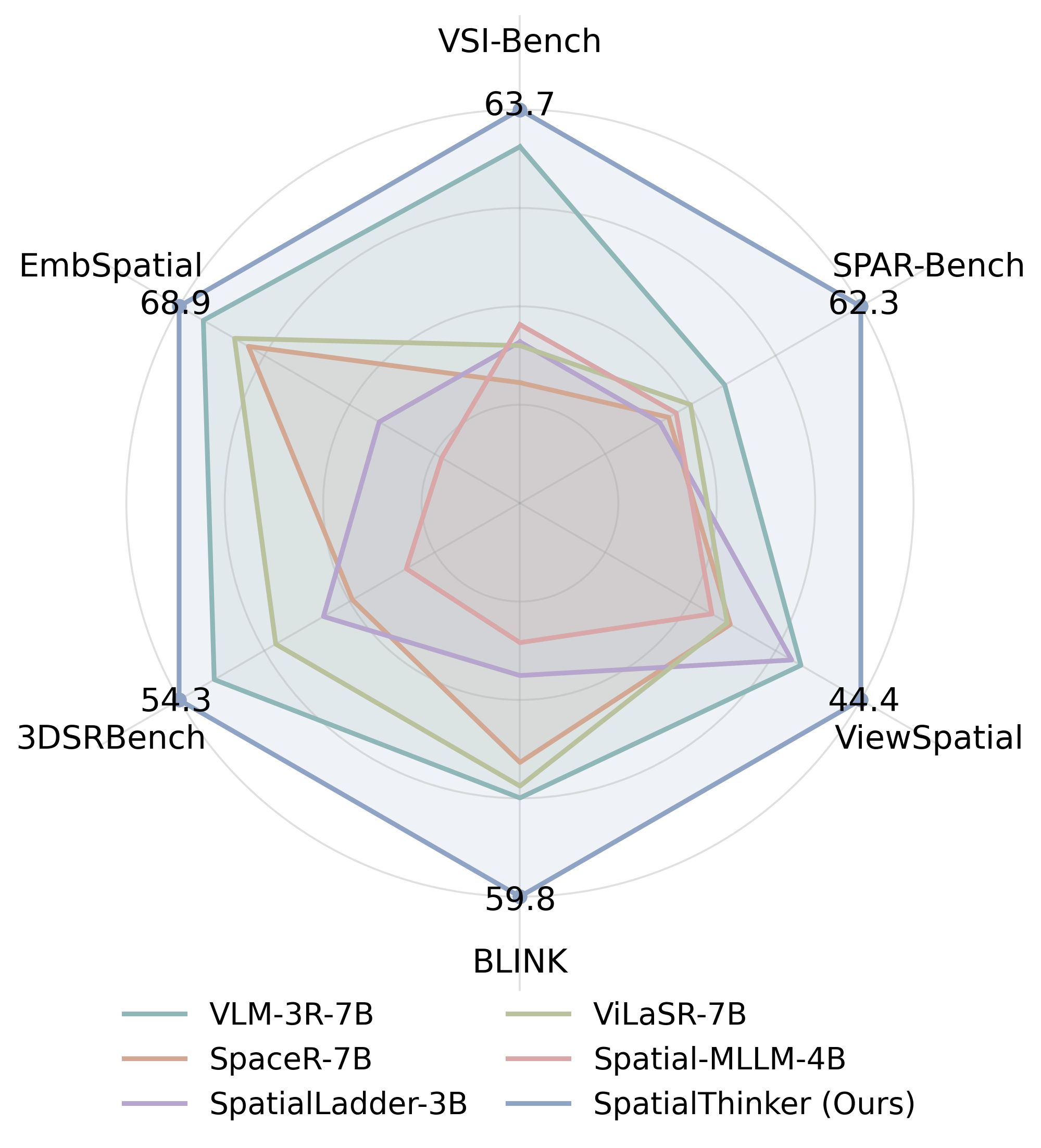}
        \label{fig:my_image}
    \end{minipage}
    \vspace{-1em}
    \caption{
    \textbf{Performance comparison on six spatial benchmarks.}
    Results include VSI-Bench~\cite{yang2025thinking}, SPAR-Bench~\cite{zhang2025flatland}, ViewSpatial~\cite{li2025viewspatial}, BLINK~\cite{fu2024blink}, 3DSRBench~\cite{ma20253dsrbench}, and EmbSpatial~\cite{du2024embspatial}.
    }
    \vspace{-1.5em}
  \label{tab:spatial-6bench}
\end{table*}

\section{Experiments}
\label{sec:exps}

\subsection{Implementation Details}

\noindent\textbf{Model and Training Setup.}
We instantiate SpatialImaginer with the BAGEL-7B-MoT~\cite{deng2025bagel} architecture. 
During the geometric-aware warm-up phase, the generation module is deactivated. We optimize the model on a 281K perception-focused subset sourced from the SPAR~\cite{zhang2025flatland}, VSI-590K~\cite{yang2025cambrian} and VLM-3R~\cite{fan2025vlm} corpora. Training utilizes the AdamW~\cite{loshchilov2017decoupled} optimizer with a cosine schedule, peaking at a learning rate of $2 \times 10^{-5}$ and decaying to a minimum of $2 \times 10^{-6}$ over 10,000 steps with 500 warmup steps. The maximum sequence length is 61,440 tokens. For visual perception, the SigLIP2~\cite{tschannen2025siglip} encoder employs a NaViT~\cite{dehghani2023patch} strategy to dynamically scale input resolutions between 224 and 518 with a patch stride of 14.

In the subsequent joint interleaved optimization phase, both visual streams are activated. The pretrained FLUX VAE~\cite{flux2024} resolution is configured to range from 256 to 512 with a stride of 16. The model is trained on our 31,503 sample balanced dataset for 4,000 steps including 200 warmup steps, with a reduced peak learning rate of $1 \times 10^{-5}$ decaying to a minimum of $1 \times 10^{-6}$. The joint objective uses equal coefficients $\lambda_{\text{text}} = 1$ and $\lambda_{\text{img}} = 1$. The rectified flow~\cite{liu2022flow} employs a timestep shift $\mu = 3.0$ alongside a visual conditioning dropout of 0.3. An exponential moving average decay of 0.995 is maintained to stabilize training across 8 NVIDIA A100.

\vspace{1mm}
\noindent\textbf{Evaluation Setup.}
We evaluate SpatialImaginer across six comprehensive spatial intelligence benchmarks: VSI-Bench~\cite{yang2025thinking}, SPAR-Bench~\cite{zhang2025flatland}, ViewSpatial~\cite{li2025viewspatial}, BLINK~\cite{fu2024blink}, 3DSRBench~\cite{ma20253dsrbench}, and EmbSpatial~\cite{du2024embspatial}. These benchmarks assess a broad spectrum of capabilities, ranging from static multi view geometry and dynamic spatiotemporal reasoning to perspective transformations and embodied spatial cognition. All evaluations are conducted under the EASI protocol~\cite{easi2025}.

\subsection{Main Results}

We compare SpatialImaginer against leading proprietary models, general-purposed open source MLLMs, and specialized spatial intelligence models across various spatial reasoning benchmarks.

\vspace{1mm}
\noindent\textbf{Results on SPAR-Bench.}
Table~\ref{tab:spar-bench} details performance on SPAR-Bench~\cite{zhang2025flatland} across 20 spatial categories stratified into low, mid, and high difficulty tiers. To enhance readability, we perform dimension reduction by averaging the scores of corresponding single-image and multi-view subtasks within each category. The complete results are provided in the appendix. SpatialImaginer achieves an overall average of 62.3. This exceeds the strongest proprietary model Seed-1.6~\cite{seed2025seed1_5vl} at 50.1 and the leading open-source spatial model VST-7B-SFT~\cite{vst2025} at 46.6. Performance stratification indicates substantial improvements on high-level tasks requiring complex geometric inference. On Object Relation tasks, our model achieved an average accuracy of 84.0 in both conditions, exceeding GPT-5~\cite{openai_gpt5_systemcard} at 64.5 and 64.3, as well as the BAGEL-7B-MoT~\cite{deng2025bagel} backbone at 51.5 and 46.7. On Spatial Imagination, SpatialImaginer reaches 58.1 and 60.3, compared to the backbone at 37.0 and 32.5. These high-level categories require explicit geometric transformations, validating that generating intermediate visual states provides tangible geometric anchors. This structural constraint effectively mitigates geometric drift during purely textual reasoning. On mid-level Position Matching tasks, SpatialImaginer achieves 84.0, outperforming VST-7B-SFT~\cite{vst2025} at 35.9 and GPT-5~\cite{openai_gpt5_systemcard} at 65.1, demonstrating that our geometric-aware warm-up training generalizes effectively across difficulty tiers.

\vspace{1mm}
\noindent\textbf{Results on VSI-Bench.}
Table~\ref{tab:vsi-bench} presents outcomes on VSI-Bench~\cite{yang2025thinking} spanning eight task categories. SpatialImaginer achieves an average of 63.7, surpassing GPT-5~\cite{openai_gpt5_systemcard} at 55.0 and open-source models including VLM-3R-7B~\cite{fan2025vlm} at 60.9 and Cambrian-S-7B~\cite{yang2025cambrian} at 62.9. The unified optimization framework improves the BAGEL-7B-MoT~\cite{deng2025bagel} backbone from an average of 31.4 to 63.7. On multiple-choice tasks, SpatialImaginer attains 69.9 in Relative Direction and 67.4 in Relative Distance, exceeding purely textual spatial models like SpaceR-7B~\cite{ouyang2025spacer} at 46.1 and 41.9, and ViLaSR-7B~\cite{wu2025reinforcing(vilasr)} at 46.5 and 45.1. For Route Planning, SpatialImaginer achieves 46.4, remaining highly competitive with GPT-5~\cite{openai_gpt5_systemcard} at 50.3. For numerical estimation, the model delivers robust scores of 70.6 for Object Count and 72.7 for Object Size. This robust performance demonstrates that explicitly generating interleaved visual states effectively grounds dynamic spatial understanding, matching or exceeding specialized systems that rely on hard-coded 3D architectural priors.

\begin{table}[t]
    \centering
    \begin{tabular}{c|ccccc}
        \toprule
        \multirow{2}{*}[-0.6ex]{Methods}  & \multicolumn{3}{c}{SPAR} & \multicolumn{2}{c}{VSI} \\
        \cmidrule(lr){2-4} \cmidrule(lr){5-6}
        & Low & Mid. & High & Num. & M.C.A. \\
        \midrule
        Baseline (SFT w/o CoT)        & 44.5 & 30.2 & 66.4 & 58.3 & 45.1 \\
        \hspace{.5em} SFT w/ Text CoT    & 42.7 & 34.8 & 61.3 & 51.7 & 47.0 \\
        \midrule
        \rowcolor{black!9}$\Delta$ & \textcolor{blue}{$\downarrow$ 1.8} & \textcolor{red}{$\uparrow$ 4.6} & \textcolor{blue}{$\downarrow$ 5.1} & \textcolor{blue}{$\downarrow$ 6.6} & \textcolor{red}{$\uparrow$ 1.9} \\
        \bottomrule
    \end{tabular}
    \caption{\textbf{Effect of textual CoT on spatial reasoning.} The two SFT variants differ solely in the inclusion of textual reasoning chains.}
    \vspace{-2.5em}
    \label{tab:ablation-textcot}
\end{table}

\begin{table}[t]
\centering
\scalebox{0.9}{
\begin{tabular}{cc|cccc}
\toprule
\textbf{GW} & \textbf{IV} & \textbf{Depth-OO} & \textbf{Cam-Pose} & \textbf{ObjRel-OC} & \textbf{SptImag-OO} \\ \midrule
        &             & 20.9              & 32.5              & 51.5                & 32.5               \\
$\checkmark$ &             & 25.6              & 28.0              & 67.3                & 46.8               \\
$\checkmark$ & $\checkmark$ & \textbf{35.2}              & \textbf{36.0}              & \textbf{84.5}               & \textbf{60.3}               \\ \bottomrule
\end{tabular}}
\caption{Performance breakdown of Geometric-aware Warm-up (GW) and Interleaved Visual CoT tuning (IV).}
\vspace{-2em}
\label{tab:ablation-interleaved}
\end{table}

\vspace{1mm}
\noindent\textbf{Cross Benchmark Evaluation.}
Table~\ref{tab:spatial-6bench} provides a comprehensive evaluation across six spatial benchmarks. SpatialImaginer achieves the highest scores on five benchmarks: VSI-Bench~\cite{yang2025thinking} at 63.7, SPAR-Bench~\cite{zhang2025flatland} at 62.3, ViewSpatial~\cite{li2025viewspatial} at 44.4, BLINK~\cite{fu2024blink} at 59.8, and 3DSRBench~\cite{ma20253dsrbench} at 54.3. Furthermore, it remains competitive on EmbSpatial~\cite{du2024embspatial} at 68.9. Previous spatial models frequently display uneven performance profiles. For instance, Cambrian-S-7B~\cite{yang2025cambrian} achieves 62.9 on VSI-Bench but drops to 37.9 on SPAR-Bench. Similarly, VLM-3R-7B~\cite{fan2025vlm} scores 60.7 on VSI-Bench but 40.5 on ViewSpatial~\cite{li2025viewspatial}. In contrast, SpatialImaginer maintains robust and uniform performance across all evaluated dimensions. This consistency demonstrates that the interleaved visual reasoning framework cultivates generalizable spatial representations rather than task specific shortcuts.

\subsection{Comprehensive Analysis}

\noindent\textbf{Effectiveness of Text-CoT.}
As discussed in Section~\ref{sec:intro}, textual CoTs qualitatively impair the model's predictions. In this section, we further provide quantitative evidence to support this observation. Building on BAGEL-7B-MoT~\cite{deng2025bagel}, we fine-tune the model using the same queries under two supervision settings: one with CoT annotations and the other with answer-only annotations. The results are presented in Table~\ref{tab:ablation-textcot}. Specifically, introducing textual CoTs leads to clear performance drops on categories that rely most heavily on precise spatial reasoning. On SPAR-Bench~\cite{zhang2025flatland}, accuracy on high-level tasks decreases from 66.4 to 61.3. On VSI-Bench~\cite{yang2025thinking}, performance on numerical estimation drops even more substantially, from 58.3 to 51.7. These categories typically involve abilities such as spatial imagination, object-relation inference, and metric estimation, all of which require the model to preserve a coherent geometric state across multiple reasoning steps. The observed degradation therefore suggests a fundamental mismatch between continuous spatial transformations and their discrete textual approximations: when spatial reasoning is forced into language-based chains, the intermediate geometric structure is more easily distorted, ultimately harming prediction accuracy.

\vspace{1mm}
\noindent\textbf{Category-wise Performance Breakdown.}
Table~\ref{tab:ablation-interleaved} summarizes the performance across different tasks. We observe that while geometric-aware warm-up effectively scales general spatial knowledge—leading to significant gains in ObjRel-OC and SptImag-OO—its impact is inconsistent across categories. Specifically, it shows limited improvement in Depth-OO and even leads to a performance degradation in Cam-Pose (32.5$\to$28.0). This suggests that scaling text-based QA alone is insufficient for tasks requiring precise geometric grounding. Crucially, interleaved visual CoT tuning serves as a vital complement; it not only recovers the performance drop in Cam-Pose (rebounding to 36.0) but also provides a substantial additive boost in Depth-OO (+9.6). These results demonstrate that visual reasoning chains are essential to bridge the gap in categories where linguistic priors fall short.

\begin{table}[t]
    \centering
    \begin{tabular}{l|ccc}
        \toprule
        Methods & SPAR & VSI & 3DSR \\
        \midrule
        Baseline (after warm-up)       & 55.6 & 59.1 & 44.6 \\
        \midrule
        \quad + ViT token         & 58.4 & 61.8 & 49.1 \\
        \quad + ViT \& VAE token  & \textbf{62.3} & \textbf{63.7} & \textbf{54.3} \\
        \bottomrule
    \end{tabular}
    \caption{\textbf{Ablation on token representation} during interleaved optimization.}
    \vspace{-2em}
    \label{tab:ablation-token}
\end{table}

\begin{table}[t]
    \centering
    \begin{tabular}{l|ccc}
        \toprule
        Rejection Rate (\%) & SPAR & VSI & VLM-3R \\
        \midrule
        Factuality Check  & 37.3 & 35.1 & 52.7 \\
        Blind Test        & 24.6 & 21.6 & 64.0 \\
        \bottomrule
    \end{tabular}
    \caption{Rejection rates of the two stage verification pipeline described in Section~\ref{sec:data_engine} across source corpora.}
    \vspace{-2em}
    \label{tab:rejection-rate}
\end{table}

\vspace{1mm}
\noindent\textbf{Necessity of VAE Tokens.}
Table~\ref{tab:ablation-token} ablates the contribution of different token representations during interleaved optimization. Training the spatially grounded foundation with only ViT tokens provides meaningful gains, improving SPAR-Bench~\cite{zhang2025flatland} by 2.8 points and VSI-Bench~\cite{yang2025thinking} by 2.7 points. This indicates that the structured sequence format alone offers inherent value. However, activating the VAE token stream to enable explicit visual generation provides substantial additional improvements, including a 3.9 point increase on SPAR-Bench and a 5.2 point increase on 3DSRBench~\cite{ma20253dsrbench}. The significant gain on 3DSRBench, which evaluates fine grained 3D spatial relationships, demonstrates that integrating continuous visual generation is crucial for establishing geometrically faithful intermediate representations.

\vspace{1mm}
\noindent\textbf{Data Quality Verification.}
Table~\ref{tab:rejection-rate} reports the rejection rates of our two stage verification pipeline. To ensure the highest quality of interleaved data, our automated engine applies exceptionally strict filtering criteria. The initial factuality check rejects 37.3 percent of SPAR~\cite{zhang2025flatland}, 35.1 percent of VSI-590K~\cite{yang2025cambrian}, and 52.7 percent of VLM-3R~\cite{fan2025vlm} candidates. The subsequent blind test further eliminates a substantial portion of the remaining data, including an additional 64.0 percent from the VLM-3R subset. These consistently high rejection rates reflect our rigorous curation process, ensuring the final dataset provides highly reliable and verifiable spatial anchors.

\section{Conclusion}

In this paper, we identify a fundamental limitation of text-only reasoning for spatial intelligence in multimodal large language models, arising from the mismatch between language-based representations and geometry-consistent reasoning. To address this, we propose SpatialImaginer, a unified multimodal framework that integrates textual reasoning with visual imagination for spatial state preservation and transformation. Experimental results demonstrate consistent improvements in robustness and state-of-the-art performance across spatial intelligence benchmarks.

\bibliographystyle{ACM-Reference-Format}
\bibliography{sample-base}

\end{document}